\documentclass[conference]{IEEEtran}
\IEEEoverridecommandlockouts
\usepackage{cite}
\usepackage{amsmath,amssymb,amsfonts}
\usepackage{algorithmic}
\usepackage{graphicx}
\usepackage{textcomp}
\usepackage{xcolor}
\usepackage{url}
\usepackage{booktabs}
\usepackage{float}
\usepackage{afterpage}
\usepackage{microtype}  % Improves spacing and hyphenation
\usepackage[english]{babel}
\def\BibTeX{{\rm B\kern-.05em{\sc i\kern-.025em b}\kern-.08em
    T\kern-.1667em\lower.7ex\hbox{E}\kern-.125emX}}
\begin{document}

\title{PaveSync: A Unified and Comprehensive Dataset for Pavement Distress Analysis and Classification\\
{}
\
}

\author{
\IEEEauthorblockN{Blessing Agyei Kyem}
\IEEEauthorblockA{\textit{Dept. of Civil, Construction, and} \\
\textit{Environmental Engineering}\\
North Dakota State University\\
Fargo, ND, USA\\
Email: blessing.agyeikyem@ndsu.edu}
\and
\IEEEauthorblockN{Joshua Kofi Asamoah}
\IEEEauthorblockA{\textit{Dept. of Civil, Construction, and} \\
\textit{Environmental Engineering}\\
North Dakota State University\\
Fargo, ND, USA\\
Email: joshua.asamoah@ndsu.edu}
\and
\IEEEauthorblockN{Anthony Dontoh}
\IEEEauthorblockA{\textit{Dept. of Civil, Construction, and} \\
\textit{Environmental Engineering}\\
The University of Memphis\\
Memphis, TN, USA\\
Email: adontoh@memphis.edu}
\and
\IEEEauthorblockN{Andrews Danyo}
\IEEEauthorblockA{\textit{Dept. of Civil, Construction, and} \\
\textit{Environmental Engineering}\\
North Dakota State University\\
Fargo, ND, USA\\
Email: andrews.danyo@ndsu.edu}
\and
\IEEEauthorblockN{Eugene Denteh}
\IEEEauthorblockA{\textit{Dept. of Civil, Construction, and} \\
\textit{Environmental Engineering}\\
North Dakota State University\\
Fargo, ND, USA\\
Email: eugene.denteh@ndsu.edu}
\and
\IEEEauthorblockN{Armstrong Aboah\textsuperscript{*}}
\IEEEauthorblockA{\textit{Dept. of Civil, Construction, and} \\
\textit{Environmental Engineering}\\
North Dakota State University\\
Fargo, ND, USA\\
Email: armstrong.aboah@ndsu.edu\\
Corresponding Author}
}

\maketitle

\begin{abstract}
Automated pavement defect detection often struggles to generalize across diverse real-world conditions due to the lack of standardized datasets. Existing datasets differ in annotation styles, distress type definitions, and formats, limiting their integration for unified training. To address this gap, we introduce a comprehensive benchmark dataset that consolidates multiple publicly available sources into a standardized collection of 52,747 images from seven countries, with 135,277 bounding box annotations covering 13 distinct distress types. The dataset captures broad real-world variation in image quality, resolution, viewing angles, and weather conditions, offering a unique resource for consistent training and evaluation. Its effectiveness was demonstrated through benchmarking with state-of-the-art object detection models (YOLOv8–YOLOv12, Faster R-CNN, and DETR), which achieved competitive performance across diverse scenarios. By standardizing class definitions and annotation formats, this dataset provides the first globally representative benchmark for pavement defect detection and enables fair comparison of models, including zero-shot transfer to new environments.
\textbf{\textit{\url{https://drive.google.com/drive/folders/1asUZbdy-3RhSVoJMLlzMOnt643ncWy6Q?usp=sharing}}}
\end{abstract}

\begin{IEEEkeywords}
Pavement distress, Deep learning, Road monitoring, Benchmarking, Datasets, Detection
\end{IEEEkeywords}

\section{Introduction}
\label{Introduction}
The integration of big data into road condition monitoring has opened new avenues for infrastructure management~\cite{shi2015big,owor2023image2pci}. Despite this progress, the road distress detection community still faces problems with fragmented datasets characterized by variations in sensor types, geographic regions, labeling schemes, and environmental conditions~\cite{ayodele2020supervised,ganin2016domain,kyem2024pavecap, kyem2024weatheradaptive}. Such fragmentation has limited the development and evaluation of deep learning models, which often excel only in narrowly defined, domain-specific settings \cite{agyeikyem2025contextcracknet, AGYEIKYEM2026106591}.

To address these challenges, several benchmark datasets such as Pavementscapes~\cite{pavementscapes}, UAV-PDD2023~\cite{yan2023uav}, and PID~\cite{majidifard2020pavement} have been introduced to improve model generalization and facilitate standardized evaluation. However, these datasets often focus on specific perspectives or limited road conditions, restricting their applicability in diverse real-world scenarios.

Building on efforts to address dataset fragmentation, this study presents \textit{PaveSync}, a large-scale, multi-perspective dataset that compiles samples from established pavement benchmarks to enhance diversity. By combining aerial, drone-based, and vehicle-mounted imagery from several countries, it captures varied environmental conditions, road surface types, and distress categories. To maintain consistency across these varied sources, a standardized annotation framework is adopted, resulting in a dataset of 52,854 annotated images spanning 13 distinct road conditions.

This standardized approach ensures a reliable basis for evaluating model performance across diverse conditions. Building on this foundation, we evaluate \textit{PaveSync} using state-of-the-art object detection models, analyzing their performance across various pavement distress types to demonstrate its effectiveness. The results highlight performance differences across various pavement distress categories, offering a fair basis for comparing different architectures. This evaluation further demonstrates how the dataset’s diversity and consistency help overcome the limitations of earlier datasets.

Beyond benchmarking, \textit{PaveSync} is designed as a resource for both researchers and practitioners in road condition monitoring. Its public availability, along with preprocessed annotations and detailed documentation, ensures accessibility and reproducibility. By providing high-quality, diverse data, this work supports the development of more robust monitoring solutions, ultimately contributing to safer roads, reduced maintenance costs, and sustainable urban development.

In summary, our contributions are as follows:
\begin{itemize}
    \item We introduce \textit{PaveSync}, a large-scale, multi-perspective dataset for road condition assessment, integrating diverse sources to improve model generalization.
    \item We establish a standardized annotation framework, ensuring consistency across datasets for reliable benchmarking of deep learning models.
    \item We benchmark several state-of-the-art object detection models on \textit{PaveSync}, providing insights into performance across various pavement distress categories.
    \item We publicly release the dataset, along with preprocessed annotations and documentation, to support reproducibility and future research. The dataset can be found via the link below: \url{https://drive.google.com/drive/folders/1asUZbdy-3RhSVoJMLlzMOnt643ncWy6Q?usp=sharing}.
\end{itemize}

The remainder of this paper is organized as follows: Section~\ref{Related Studies} reviews related studies on pavement defect detection, Section~\ref{data} describes the datasets used, and Section~\ref{method} discusses the various models used to benchmark PaveSync. In Section~\ref{results}, we present the results of the object detection models and explore potential use cases of our dataset. Finally, Section~\ref{conclusion} concludes the work.

\section{Related Studies}
\label{Related Studies}

In automated pavement distress detection, numerous datasets have been developed to train and evaluate deep learning models. The GAPs dataset by Eisenbach et al.~\cite{eisenbach2017get} was the first large-scale, freely available dataset, featuring 1,969 grayscale images from German federal roads but with limited climate diversity. 

Subsequent datasets addressed these limitations through multi-perspective imagery and broader environmental conditions. Majidifard et al. introduced PID~\cite{majidifard2020pavement} with 7,237 manually annotated images across seven distress types, while the ISTD-PDS7 dataset~\cite{song2023istd} provided 18,527 images across nine scenarios. However, both lack coverage of extreme weather conditions. Liu et al. developed PaveDistress~\cite{liu2024pavedistress}, offering high-resolution imagery with fine-grained labels under varying lighting conditions, though limited to a single highway system.

The UAV-PDD2023 dataset~\cite{yan2023uav} leveraged unmanned aerial vehicles, capturing over 11,150 images with diverse weather and sensor conditions, though UAV-based approaches face occlusion and distortion challenges. For segmentation tasks, Pavementscapes~\cite{pavementscapes} provided 4,000 high-resolution images across 15 pavement types with 8,680 labeled damage instances. The geographically diverse RDD2022~\cite{RDD2022} dataset spans multiple countries, while DSPS~\cite{DSPS} offered benchmarked data for algorithm development competitions.

Despite these advances, significant limitations persist: datasets typically focus on single sensor types, lack extreme environmental variations, and suffer from annotation inconsistencies that hinder effective benchmarking. \textit{PaveSync} addresses these gaps by providing a diverse, large-scale dataset with multi-perspective imagery, standardized annotations, and comprehensive environmental conditions to enhance model generalization and enable fair benchmarking across deep learning architectures.

\section{Dataset}
\label{data}
\subsection{Data Sources and Composition}
\begin{figure}[t!]
  \centering
  \includegraphics[width=0.45\textwidth]{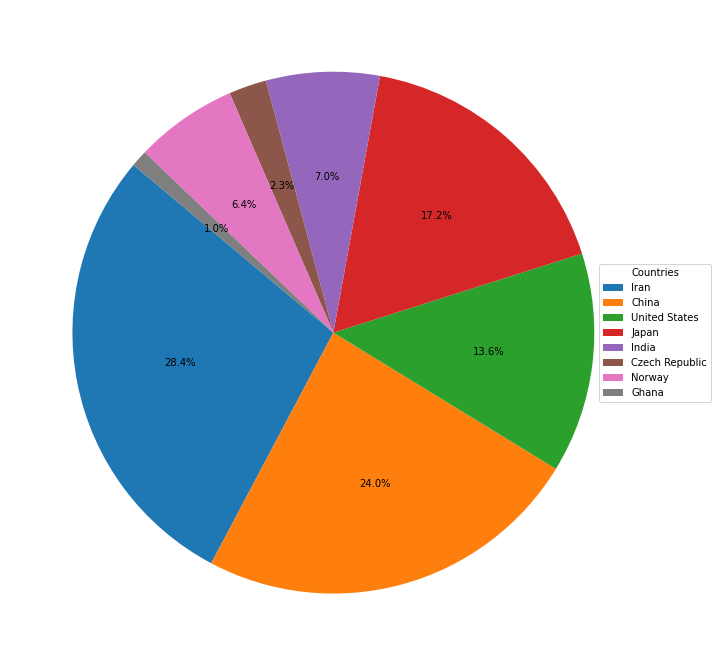}
  \caption{Distribution of the PaveSync dataset across different countries}
  \label{fig:country}
\end{figure}

This study introduces a newly compiled dataset of pavement distress images aggregated from multiple publicly available sources \cite{pavementscapes, UAPD, liu2024pavedistress, DSPS, owor2024pavesam, RDD2022, kyem2024advancing, kyem2024pavecap}. The data originates from diverse geographic regions, including Iran, China, the United States, and Ghana, as illustrated in Figure \ref{fig:country}.

\begin{figure}[t!]
  \centering
  \includegraphics[width=0.48\textwidth]{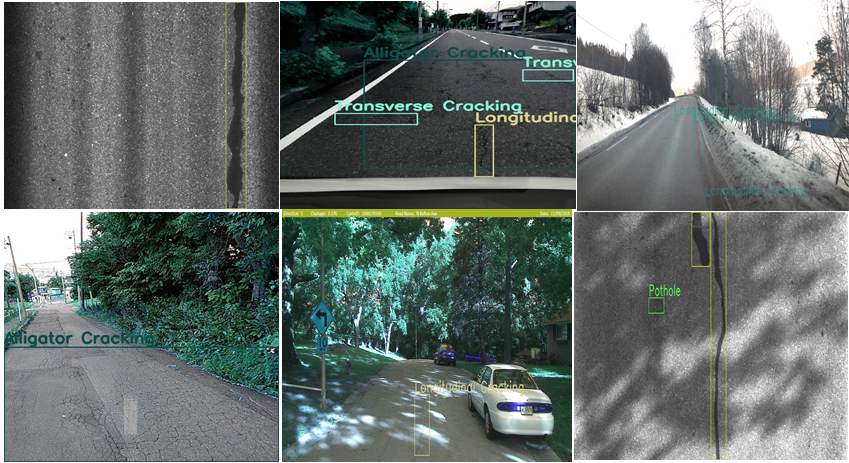}
  \caption{Different imaging orientations in the dataset, including ground-level, aerial, and top-down views}
  \label{fig:distress_images_by_country}
\end{figure}

Beyond geographic diversity, PaveSync incorporates multiple orientations and perspectives, ensuring comprehensive coverage of pavement distress types under varying imaging conditions. As illustrated in Figure \ref{fig:distress_images_by_country}, the dataset includes ground-level, pavement-level, aerial (drone), and top-down views, each offering unique advantages for distress detection. These variations ensure that the dataset is not biased toward a specific imaging condition, making it more adaptable, and generalizable across real-world pavement monitoring scenarios. 

\begin{figure}[t!]
  \centering
  \includegraphics[width=0.48\textwidth]{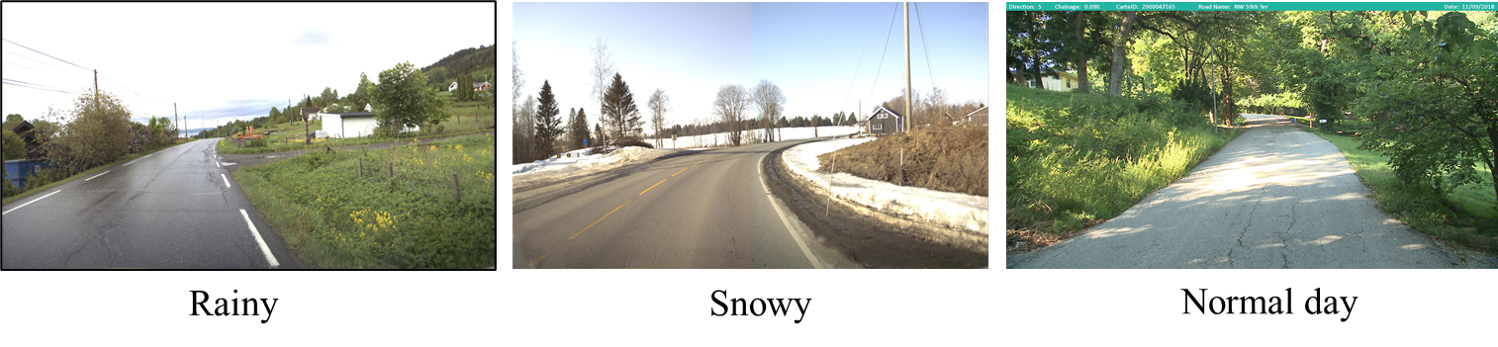}
  \caption{Sample images from the dataset captured under different weather conditions}
  \label{weather_conditions}
\end{figure}

In addition to diverse orientations and perspectives, the dataset also captures images under varying weather and environmental conditions. It includes images collected during daylight, snowy, and rainy conditions, representing real-world scenarios where pavement distress detection systems must operate reliably, as shown in Figure \ref{weather_conditions}. Incorporating diverse weather conditions strengthens the dataset’s ability to handle real-world variability in pavement monitoring.

\subsection{Data Standardization}
The datasets utilized in this study had varying annotation formats such as Pascal VOC (XML), COCO (JSON), and YOLO (TXT), each with distinct structures for representing bounding boxes and object metadata. While some formats provide detailed annotations, including segmentation masks and hierarchical labels, others focus on minimal yet efficient representations for real-time applications.

To create a unified dataset, we standardized the annotation formats, distress class names, and class IDs to ensure consistency and compatibility across different sources. Rather than restricting the dataset to a single annotation style, we retained multiple formats—XML, JSON, and TXT (YOLO)—allowing researchers to work with their preferred structure. This approach ensures flexibility while maintaining a standardized class mapping across formats.

Additionally, distress class names varied across datasets, introducing inconsistencies in labeling. For instance, "Alligator" in one dataset appeared as "Alligator Cracking" in another. To address this, we assigned standardized distress names, removed duplicates, and eliminated ambiguous labels and unwanted distress types. Another challenge was the inconsistency in class ID assignments, where different datasets used different IDs for the same distress type. We standardized these identifiers, ensuring that each distress type had a unique and consistent class ID across the final dataset. Furthermore, distress classes with an insufficient number of samples were removed to mitigate dataset imbalance and improve training stability.

After merging and standardizing all datasets, the final set of distress classes is presented in Table \ref{tab:dataset_summary}, ensuring a structured and consistent dataset that remains versatile for various research applications and annotation preferences.

\begin{table}[H]
    \centering
    \caption{Distribution of dataset across countries and distress types in training and validation splits.}
    \label{tab:dataset_summary}
    \begin{tabular}{c p{3.0cm} p{0.9cm} p{0.9cm} p{1cm}}
        \toprule
        \textbf{Country No} & \textbf{Country} & \textbf{Train} & \textbf{Val} & \textbf{Total} \\
        \midrule
        0 & Iran            & 13485 & 1498 & 14983 \\
        1 & China           & 11394 & 1266 & 12660 \\
        2 & United States   & 6457  & 717  & 7174  \\
        3 & Japan           & 8164  & 907  & 9071  \\
        4 & India           & 3323  & 369  & 3692  \\
        5 & Czech Republic  & 1092  & 121  & 1213  \\
        6 & Norway          & 3038  & 338  & 3376  \\
        7 & Ghana           & 520   & 58   & 578   \\
        \midrule
        \textbf{Class ID} & \textbf{Distress Type} & \textbf{Train} & \textbf{Val} & \textbf{Total} \\
        \midrule
        0  & Bleeding              & 1690  & 195   & 1885  \\
        1  & Bumps and Sags        & 784   & 73    & 857   \\
        2  & Manhole               & 721   & 75    & 796   \\
        3  & Patching              & 4629  & 492   & 5121  \\
        4  & Pothole               & 25874 & 2764  & 28638 \\
        5  & Rutting               & 15456 & 1943  & 17399 \\
        6  & Shoving               & 1394  & 162   & 1556  \\
        7  & Alligator Cracking    & 18635 & 2042  & 20677 \\
        8  & Longitudinal Cracking & 29969 & 3384  & 33353 \\
        9  & Transverse Cracking   & 17520 & 1931  & 19451 \\
        10 & Block Cracking        & 402   & 44    & 446   \\
        11 & Repair                & 2995  & 705   & 3700  \\
        12 & Edge Cracking         & 1523  & 191   & 1714  \\
        \midrule
           & \textbf{Total Images}   & 47473 & 5274 & 52747 \\
           & \textbf{Bounding Boxes} & --    & --   & 135277 \\
        \bottomrule
    \end{tabular}
\end{table}

\subsection{Annotation Validation}
Due to the large volume of standardized images, a full manual review of all annotations was impractical. To maintain accuracy and consistency, a stratified sampling-based validation strategy was implemented. This approach involved selecting a representative subset of images, ensuring coverage across various distress types, environmental conditions, and geographic locations. The validation process included visually overlaying both the standardized and original annotations onto the corresponding images to assess their alignment. Any discrepancies identified were corrected and re-evaluated, with the process repeated iteratively until all annotations accurately reflected the actual pavement conditions. This structured methodology ensured high-quality ground-truth labeling across all datasets integrated into the benchmark.

\subsection{Final Dataset Composition}
The final combined dataset consists of 52,747 pavement images across 13 distress classes (see Table \ref{tab:dataset_summary}) with a total of 135,277 bounding box annotations. The dataset was then split into 90\% training and 10\% validation, ensuring that each distress class maintained the same distribution across both sets. This stratified split prevents class imbalance during training and evaluation.  To enhance generalization, we ensured a balanced mix of images from different geographic regions.

\section{Methodology}
\label{method}
This section describes the methodology used to preprocess, benchmark, and evaluate PaveSync against existing deep learning models. The process includes data preprocessing, training and evaluation setup, evaluation metrics, and benchmarking with multiple architectures. 

\subsection{Data Preprocessing and Augmentation}
\label{preprocessing}
To ensure consistency and enhance model performance, several preprocessing steps were applied to the PaveSync dataset. All images were resized to a standard dimension of 640 × 640 pixels while maintaining aspect ratios to ensure compatibility with different deep learning models. In addition, various data augmentation techniques, including random cropping, rotation, flipping, brightness adjustments, and Gaussian noise, were implemented to improve model generalization across varying real-world conditions. Finally, since certain distress types appeared more frequently than others, weighted sampling was applied during training to address class imbalance and ensure models learned effectively. The specific preprocessing and augmentation techniques used are summarized in Table \ref{tab:preprocessing}.

\begin{table}[!ht]
    \centering
    \caption{Preprocessing and Augmentation Settings for PaveSync}
    \begin{tabular}{lc}
        \toprule
        \textbf{Technique} & \textbf{Value} \\ 
        \midrule
        Random Cropping & 0.8 \\ 
        Rotation (degrees) & 15 \\ 
        Horizontal Flipping (\%) & 0.5 \\ 
        Brightness Adjustment & 1.1 \\ 
        Contrast Adjustment & 1.2 \\ 
        Gaussian Noise Std & 0.01 \\ 
        Normalization (Pixel Scale) & [0,1] \\ 
        \bottomrule
    \end{tabular}
    \label{tab:preprocessing}
\end{table}

\subsection{Experimental Setup}
The dataset was divided into two subsets to ensure robust model training and evaluation: 90\% for training, and 10\% for validation. This split ensures sufficient training data while preserving an independent test set for final evaluation. All experiments were conducted on an NVIDIA A100 GPU with 24 GB VRAM, using PyTorch 2.0 as the deep learning framework. The models were trained for 1000 epochs with a batch size of 16, using the Adam optimizer with a learning rate of 0.001, decayed using a cosine annealing schedule.

\subsection{Deep Learning Models for Benchmarking}
To comprehensively evaluate the PaveSync dataset, we benchmarked it using 7 state-of-the-art models, each designed to handle different trade-offs between accuracy, computational efficiency, and real-world applicability. These models have been widely used in the field of pavement distress detection, road surface analysis, and general object detection tasks.

\subsubsection{YOLO Variants}
The YOLO family of detectors (v8–v12)
\cite{aboah2023real,danyo2025resnet50pci,yolo11_ultralytics,tian2025yolov12attentioncentricrealtimeobject}
were selected for benchmarking due to their balance of speed and accuracy in real-time object detection. All employ anchor-free architectures suited for dense pavement imagery, but each introduces targeted refinements relevant to pavement distress detection. YOLOv8 \cite{aboah2023real}, the baseline, integrates an anchor-free detection head, dynamic label assignment, and a CSPDarkNet backbone with a decoupled head, providing strong baseline accuracy and efficiency. YOLOv9 \cite{wang2024yolov9} improves feature extraction through programmable gradient backpropagation and GELAN, enhancing detection of fine-grained cracks. YOLOv10 \cite{wang2024yolov10realtimeendtoendobject} eliminates redundant post-processing through an NMS-free design and employs dual assignments, improving consistency and efficiency for real-time road monitoring systems. YOLOv11 \cite{yolo11_ultralytics} strengthens multi-scale feature aggregation with the C3k2 block, SPPF, and spatial attention modules, which improve robustness to scale variation among defects such as potholes versus block cracks. Finally, YOLOv12 \cite{tian2025yolov12attentioncentricrealtimeobject} introduces attention-centric modules, separable convolutions, and FlashAttention, improving generalization across varying image qualities and conditions typical of multi-country pavement datasets. These incremental improvements directly address challenges of defect variability in scale, texture, and capture environments without repeating baseline YOLO characteristics.

\subsubsection{Faster R-CNN}
Faster R-CNN \cite{ren2016fasterrcnnrealtimeobject} is a two-stage object detection model known for its high detection accuracy, particularly for small or densely packed objects \cite{dontoh2025visualdominanceemergingmultimodal}. It consists of a Region Proposal Network (RPN) followed by a classification and regression network, allowing it to refine bounding box predictions iteratively. The model uses ResNet-50 as the feature extractor, making it highly effective for detecting fine-grained pavement defects such as microcracks and texture inconsistencies.

\subsubsection{DETR: Detection Transformers}
Detection Transformer (DETR) \cite{carion2020end} employs a self-attention mechanism for object detection. This architecture removes the need for 
non-maximum suppression (NMS) and enables direct set predictions, making it highly effective for complex road textures and overlapping pavement defects. 

\begin{table*}[t!]
\centering
\caption{Detection performance of YOLOv8–YOLOv12, Faster R-CNN, and DETR across 13 distinct pavement distresses in our standardized benchmark dataset. Each sub-table reports Precision, Recall, mAP@50, and mAP@50–95 per class}
\label{tab:combined_results}

%----------------- Part 1: Bleeding, Bumps & Sags, Manhole, Patching -------------
% \textbf{(Part 1) Bleeding, Bumps \& Sags, Manhole, and Patching}\\[6pt]
\resizebox{\textwidth}{!}{%
\begin{tabular}{l
cccc % Bleeding
cccc % Bumps & Sags
cccc % Manhole
cccc % Patching
}
\toprule
& \multicolumn{4}{c}{\textbf{Bleeding}}
& \multicolumn{4}{c}{\textbf{Bumps \& Sags}}
& \multicolumn{4}{c}{\textbf{Manhole}}
& \multicolumn{4}{c}{\textbf{Patching}} \\
\cmidrule(lr){2-5} \cmidrule(lr){6-9} \cmidrule(lr){10-13} \cmidrule(lr){14-17}
\textbf{Model}
& \textbf{Prec} & \textbf{Rec} & \textbf{mAP@50} & \textbf{mAP@50-95}
& \textbf{Prec} & \textbf{Rec} & \textbf{mAP@50} & \textbf{mAP@50-95}
& \textbf{Prec} & \textbf{Rec} & \textbf{mAP@50} & \textbf{mAP@50-95}
& \textbf{Prec} & \textbf{Rec} & \textbf{mAP@50} & \textbf{mAP@50-95} \\
\midrule

%% YOLOv8
\textbf{YOLOv8}
& 0.448 & 0.441 & 0.367 & 0.213
& 0.473 & 0.356 & 0.379 & 0.194
& 0.616 & 0.920 & 0.834 & 0.602
& 0.603 & 0.682 & 0.690 & 0.530
\\

%% YOLOv9
\textbf{YOLOv9}
& 0.433 & 0.379 & 0.336 & 0.196
& 0.548 & 0.315 & 0.413 & 0.222
& 0.690 & 0.920 & 0.860 & 0.620
& 0.611 & 0.662 & 0.666 & 0.512
\\

%% YOLOv10
\textbf{YOLOv10}
& 0.406 & 0.354 & 0.310 & 0.185
& 0.571 & 0.329 & 0.453 & 0.217
& 0.663 & 0.893 & 0.824 & 0.575
& 0.617 & 0.648 & 0.668 & 0.512
\\

%% YOLOv11
\textbf{YOLOv11}
& 0.425 & 0.262 & 0.296 & 0.197
& 0.512 & 0.288 & 0.407 & 0.205
& 0.625 & 0.867 & 0.809 & 0.573
& 0.640 & 0.658 & 0.678 & 0.522
\\

%% YOLOv12
\textbf{YOLOv12}
& 0.429 & 0.374 & 0.341 & 0.201
& 0.541 & 0.274 & 0.405 & 0.202
& 0.674 & 0.827 & 0.806 & 0.569
& 0.610 & 0.648 & 0.643 & 0.489
\\

%% Faster-RCNN
\textbf{Faster R-CNN}
& 0.460 & 0.450 & 0.365 & 0.210
& 0.520 & 0.310 & 0.410 & 0.210
& 0.640 & 0.890 & 0.835 & 0.590
& 0.620 & 0.650 & 0.670 & 0.515
\\

%% DETR
\textbf{DETR}
& 0.450 & 0.400 & 0.350 & 0.195
& 0.520 & 0.280 & 0.405 & 0.200
& 0.620 & 0.880 & 0.830 & 0.560
& 0.610 & 0.640 & 0.655 & 0.495
\\

\bottomrule
\end{tabular}
}
% End Part 1

\vspace{1.5em}
%----------------- Part 2: Pothole, Rutting, Shoving, Alligator Cracking -------------
% \textbf{(Part 2) Pothole, Rutting, Shoving, and Alligator Cracking}\\[6pt]
\resizebox{\textwidth}{!}{%
\begin{tabular}{l
cccc % Pothole
cccc % Rutting
cccc % Shoving
cccc % Alligator
}
\toprule
& \multicolumn{4}{c}{\textbf{Pothole}}
& \multicolumn{4}{c}{\textbf{Rutting}}
& \multicolumn{4}{c}{\textbf{Shoving}}
& \multicolumn{4}{c}{\textbf{Alligator}} \\
\cmidrule(lr){2-5} \cmidrule(lr){6-9} \cmidrule(lr){10-13} \cmidrule(lr){14-17}
\textbf{Model}
& \textbf{Prec} & \textbf{Rec} & \textbf{mAP@50} & \textbf{mAP@50-95}
& \textbf{Prec} & \textbf{Rec} & \textbf{mAP@50} & \textbf{mAP@50-95}
& \textbf{Prec} & \textbf{Rec} & \textbf{mAP@50} & \textbf{mAP@50-95}
& \textbf{Prec} & \textbf{Rec} & \textbf{mAP@50} & \textbf{mAP@50-95} \\
\midrule

%% YOLOv8
\textbf{YOLOv8}
& 0.731 & 0.737 & 0.790 & 0.519
& 0.744 & 0.952 & 0.916 & 0.768
& 0.778 & 0.778 & 0.802 & 0.423
& 0.680 & 0.726 & 0.759 & 0.543
\\

%% YOLOv9
\textbf{YOLOv9}
& 0.743 & 0.693 & 0.761 & 0.474
& 0.733 & 0.949 & 0.912 & 0.758
& 0.776 & 0.815 & 0.810 & 0.429
& 0.708 & 0.691 & 0.746 & 0.526
\\

%% YOLOv10
\textbf{YOLOv10}
& 0.711 & 0.696 & 0.759 & 0.475
& 0.747 & 0.937 & 0.910 & 0.759
& 0.677 & 0.778 & 0.760 & 0.493
& 0.713 & 0.684 & 0.741 & 0.525
\\

%% YOLOv11
\textbf{YOLOv11}
& 0.765 & 0.682 & 0.761 & 0.470
& 0.751 & 0.940 & 0.986 & 0.756
& 0.749 & 0.790 & 0.791 & 0.419
& 0.702 & 0.693 & 0.743 & 0.527
\\

%% YOLOv12
\textbf{YOLOv12}
& 0.746 & 0.648 & 0.732 & 0.431
& 0.736 & 0.942 & 0.905 & 0.747
& 0.724 & 0.778 & 0.778 & 0.408
& 0.712 & 0.660 & 0.726 & 0.516
\\

%% Faster-RCNN
\textbf{Faster R-CNN}
& 0.730 & 0.700 & 0.760 & 0.500
& 0.740 & 0.930 & 0.920 & 0.760
& 0.710 & 0.780 & 0.785 & 0.460
& 0.695 & 0.680 & 0.750 & 0.530
\\

%% DETR
\textbf{DETR}
& 0.720 & 0.660 & 0.740 & 0.460
& 0.740 & 0.940 & 0.900 & 0.750
& 0.700 & 0.770 & 0.780 & 0.420
& 0.710 & 0.680 & 0.740 & 0.520
\\

\bottomrule
\end{tabular}
}
% End Part 2

\vspace{1.5em}
%----------------- Part 3: Longitudinal, Transverse, Block, Repair, Edge -------------
% \textbf{(Part 3) Longitudinal Cracking, Transverse Cracking, Block Cracking, Repair, and Edge Cracking}\\[6pt]
\resizebox{\textwidth}{!}{%
\begin{tabular}{l
cccc % Longitudinal
cccc % Transverse
cccc % Block
cccc % Repair
cccc % Edge
}
\toprule
& \multicolumn{4}{c}{\textbf{Longitudinal}}
& \multicolumn{4}{c}{\textbf{Transverse}}
& \multicolumn{4}{c}{\textbf{Block Cracking}}
& \multicolumn{4}{c}{\textbf{Repair}}
& \multicolumn{4}{c}{\textbf{Edge Cracking}} \\
\cmidrule(lr){2-5} \cmidrule(lr){6-9} \cmidrule(lr){10-13}
\cmidrule(lr){14-17} \cmidrule(lr){18-21}
\textbf{Model}
& \textbf{P} & \textbf{R} & \textbf{m@50} & \textbf{m@50-95}
& \textbf{P} & \textbf{R} & \textbf{m@50} & \textbf{m@50-95}
& \textbf{P} & \textbf{R} & \textbf{m@50} & \textbf{m@50-95}
& \textbf{P} & \textbf{R} & \textbf{m@50} & \textbf{m@50-95}
& \textbf{P} & \textbf{R} & \textbf{m@50} & \textbf{m@50-95} \\
\midrule

%% YOLOv8
\textbf{YOLOv8}
& 0.613 & 0.615 & 0.624 & 0.370
& 0.608 & 0.664 & 0.659 & 0.381
& 0.553 & 0.605 & 0.614 & 0.469
& 0.690 & 0.860 & 0.856 & 0.759
& 0.754 & 0.770 & 0.825 & 0.470
\\

%% YOLOv9
\textbf{YOLOv9}
& 0.611 & 0.570 & 0.597 & 0.354
& 0.623 & 0.615 & 0.642 & 0.364
& 0.556 & 0.581 & 0.603 & 0.443
& 0.703 & 0.854 & 0.856 & 0.755
& 0.692 & 0.717 & 0.776 & 0.415
\\

%% YOLOv10
\textbf{YOLOv10}
& 0.616 & 0.533 & 0.584 & 0.347
& 0.626 & 0.586 & 0.638 & 0.359
& 0.676 & 0.535 & 0.633 & 0.457
& 0.728 & 0.820 & 0.841 & 0.748
& 0.687 & 0.654 & 0.730 & 0.410
\\

%% YOLOv11
\textbf{YOLOv11}
& 0.628 & 0.549 & 0.595 & 0.357
& 0.621 & 0.610 & 0.641 & 0.394
& 0.619 & 0.605 & 0.615 & 0.471
& 0.713 & 0.849 & 0.856 & 0.758
& 0.739 & 0.681 & 0.779 & 0.414
\\

%% YOLOv12
\textbf{YOLOv12}
& 0.659 & 0.512 & 0.587 & 0.345
& 0.632 & 0.573 & 0.625 & 0.345
& 0.619 & 0.605 & 0.628 & 0.475
& 0.682 & 0.850 & 0.851 & 0.752
& 0.691 & 0.634 & 0.728 & 0.370
\\

%% Faster-RCNN
\textbf{Faster-RCNN}
& 0.620 & 0.580 & 0.610 & 0.360
& 0.615 & 0.600 & 0.640 & 0.380
& 0.610 & 0.570 & 0.620 & 0.450
& 0.695 & 0.840 & 0.850 & 0.752
& 0.720 & 0.680 & 0.760 & 0.420
\\

%% DETR
\textbf{DETR}
& 0.600 & 0.540 & 0.590 & 0.350
& 0.590 & 0.600 & 0.635 & 0.360
& 0.620 & 0.590 & 0.630 & 0.440
& 0.680 & 0.830 & 0.850 & 0.740
& 0.720 & 0.660 & 0.750 & 0.410
\\

\bottomrule
\end{tabular}
}
% End Part 3

\end{table*}

\subsection{Evaluation Metrics}
To ensure a fair and standardized comparison of deep learning models for pavement distress detection, we employ a set of widely used evaluation metrics. 

\subsubsection{Precision (\(P\))}
Precision measures the proportion of correctly predicted pavement distress instances among all detected instances. It is defined as:

\begin{equation}
P = \frac{TP}{TP + FP}
\end{equation}

where \( TP \) (True Positives) represents correctly identified distress instances, and \( FP \) (False Positives) denotes incorrectly detected instances. A higher precision value indicates fewer false alarms, making it critical for applications requiring high detection reliability.

\subsubsection{Recall (\(R\))}
Recall evaluates how well the model detects actual pavement distress instances in the dataset. It is computed as:

\begin{equation}
R = \frac{TP}{TP + FN}
\end{equation}

where \( FN \) (False Negatives) represents missed distress instances. High recall ensures that most pavement defects are correctly identified, which is crucial for infrastructure assessment applications.

\subsubsection{F1-Score}
The F1-score provides a balance between precision and recall, ensuring that neither false positives nor false negatives dominate the evaluation. It is calculated as:

\begin{equation}
F1 = 2 \times \frac{P \times R}{P + R}
\end{equation}

A high F1-score indicates that the model performs well in both correctly detecting defects and minimizing false detections.

\subsubsection{Mean Average Precision}
Mean Average Precision (mAP) evaluates object detection accuracy by averaging the precision values across multiple Intersection over Union (IoU) thresholds. 

\begin{itemize}
    \item \textbf{mAP@50}: Average Precision at a fixed IoU threshold of 50\%.
    \item \textbf{mAP@50-95}: Average Precision computed across IoU thresholds from 50\% to 95\% in 5\% increments, ensuring stricter evaluation.
\end{itemize}

It is defined as:

\begin{equation}
mAP = \frac{1}{N} \sum_{i=1}^{N} AP_i
\end{equation}

where \(N\) is the number of IoU thresholds, and \(AP_i\) is the Average Precision at each threshold. A higher mAP value indicates better model performance.

\section{Results and Use Case}
\label{results}
\subsection{Results}
The results in Table \ref{tab:combined_results} show that each YOLO variant excels in certain distress classes while performing less consistently on others. YOLOv8 offers strong results for rutting and shoving, suggesting its feature extraction is well-tuned for large, easily recognizable deformations. YOLOv9 stands out for higher recall on manhole and rutting, indicating robust ability to capture these anomalies even when less obvious. YOLOv10 balances precision and recall across most classes, suggesting its architectural refinements enhance detection consistency. YOLOv11 achieves particularly high precision in complex categories like alligator cracking, likely due to its focus on localizing intricate features. YOLOv12 maintains solid overall performance across many defects but struggles with recall on irregular classes, hinting at difficulties in learning subtle patterns.

Beyond the YOLO family, Faster R-CNN and DETR display competitive detection outcomes in manhole and rutting, reflecting these models' capacity to model context. Faster R-CNN's region-based approach provides balanced performance across various defects, while DETR's attention-driven architecture leads to high recall for classes with distinct shape signatures, though it exhibits slightly lower precision on subtler problems such as bumps and sags. These findings emphasize how each model's design influences its sensitivity and specificity for different pavement distress types, offering opportunities to refine models for more accurate detection in future work.

% \afterpage{
% \clearpage

\subsection{Use Cases of the Dataset}
The dataset serves multiple pivotal functions within both research and real-world operational contexts, enabling advancements in pavement monitoring, infrastructure maintenance, and intelligent transportation systems. First, its large and diversified image repository facilitates the training, validation, and benchmarking of deep learning models, thereby helping researchers and practitioners enhance accuracy across varied distress types and environmental conditions. In parallel, transportation agencies can integrate these models into real-time monitoring pipelines to detect critical defects, such as potholes or severe cracking, and prioritize repairs before they worsen. Because the dataset encompasses imagery collected from multiple geographic locations and under different weather conditions, it supports domain adaptation and transfer learning studies. This makes it particularly valuable for scaling object detection systems to new regions or sensor modalities. Furthermore, city planners and maintenance contractors can exploit the standardized format and comprehensive labeling to develop predictive maintenance schedules, streamlining resource allocation and optimizing long-term infrastructure investments. In educational settings, the dataset offers a rigorous testbed for hands-on projects, allowing students to learn state-of-the-art object detection techniques and gain practical insights into road asset management. 

\section{Conclusion}
\label{conclusion}
In conclusion, PaveSync brings together pavement images from multiple perspectives, climates, and sensor types into a single, standardized resource that addresses the fragmented nature of existing datasets. By harmonizing annotation formats and unifying class definitions, the dataset enables fair and consistent benchmarking across diverse real-world conditions. Baseline comparisons with different architectures highlight PaveSync’s ability to reveal each model’s strengths and shortcomings, underscoring its value as a comprehensive evaluation tool. With global coverage and consistent labeling, the dataset lays the groundwork for new research in automated road condition assessment, supporting the development of foundation models that combine deep learning algorithms and diverse data inputs to improve prediction accuracy and operational efficiency. In this way, PaveSync not only connects scattered data sources but also inspires new strategies for smarter, more reliable road maintenance in the future.

\section{Acknowledgements}
This research was supported by the North Dakota Economic Diversification Research Fund (EDRF).The author(s) express gratitude for the funding and support that made this work possible.


\begin{thebibliography}{00}

\bibitem{shi2015big}
Q.~Shi and M.~Abdel-Aty, ``Big data applications in real-time traffic operation and safety monitoring and improvement on urban expressways,'' \emph{Transp. Res. Part C Emerg. Technol.}, vol.~58, pp. 380--394, Sep. 2015.

\bibitem{owor2023image2pci}
N.~J. Owor, Y.~Adu-Gyamfi, and M.~Amo-Boateng, ``Image2PCI: Vision transformer with multi-task learning for automated pavement condition index estimation,'' \emph{IEEE Access}, vol.~11, pp. 121\,894--121\,909, 2023.

\bibitem{ayodele2020supervised}
K.~P. Ayodele, W.~O. Ikezogwo, M.~A. Komolafe, and P.~Ogunbona, ``Supervised domain generalization for integration of disparate scalp EEG datasets for automatic epileptic seizure detection,'' \emph{Comput. Biol. Med.}, vol. 120, p. 103757, Mar. 2020.

\bibitem{ganin2016domain}
Y.~Ganin \emph{et al.}, ``Domain-adversarial training of neural networks,'' \emph{J. Mach. Learn. Res.}, vol.~17, no.~59, pp. 1--35, 2016.

\bibitem{kyem2024pavecap}
B.~A. Kyem \emph{et al.}, ``Pavecap: The first multimodal framework for comprehensive pavement condition assessment with dense captioning and PCI estimation,'' \emph{arXiv preprint arXiv:2408.04110}, 2024.

\bibitem{kyem2024weatheradaptive}
B.~A. Kyem, J.~K. Asamoah, Y.~Huang, and A.~Aboah, ``Weather-adaptive synthetic data generation for enhanced power line inspection using StarGAN,'' \emph{IEEE Access}, vol.~12, pp. 193882--193901, 2024, doi: 10.1109/ACCESS.2024.3520120.

\bibitem{agyeikyem2025contextcracknet}
B.~A. Kyem, J.~K. Asamoah, and A.~Aboah, ``Context-CrackNet: A context-aware framework for precise segmentation of tiny cracks in pavement images,'' \emph{Construction and Building Materials}, vol.~484, p. 141583, 2025, doi: 10.1016/j.conbuildmat.2025.141583. [Online]. Available: https://www.sciencedirect.com/science/article/pii/S0950061825017337

\bibitem{AGYEIKYEM2026106591}
B.~Agyei Kyem \emph{et al.}, ``Self-supervised multi-scale transformer with Attention-Guided Fusion for efficient crack detection,'' \emph{Autom. Constr.}, vol. 181, p. 106591, 2026.

\bibitem{pavementscapes}
Z.~Tong, T.~Ma, J.~Huyan, and W.~Zhang, ``Pavementscapes: A large-scale hierarchical image dataset for asphalt pavement damage segmentation,'' \emph{arXiv preprint arXiv:2208.00775}, 2022.

\bibitem{yan2023uav}
K.~Yan \emph{et al.}, ``UAV-PDD2023: A high-resolution UAV pavement distress detection dataset,'' in \emph{Proc. IEEE/CVF Conf. Comput. Vis. Pattern Recognit. Workshops}, 2023, pp. 1--9.

\bibitem{majidifard2020pavement}
H.~Majidifard, M.~J. Buttlar, and H.~Alavi, ``Pavement image dataset (PID): A new benchmark dataset for pavement distress detection,'' \emph{Data Brief}, vol.~31, p. 105961, Aug. 2020.

\bibitem{eisenbach2017get}
M.~Eisenbach \emph{et al.}, ``How to get pavement distress detection ready for deep learning? A systematic approach,'' in \emph{Proc. Int. Joint Conf. Neural Netw. (IJCNN)}, 2017, pp. 2039--2047.

\bibitem{song2023istd}
Y.~Song \emph{et al.}, ``ISTD-PDS7: An image dataset for pavement distress segmentation in seven scenarios,'' \emph{Data Brief}, vol.~48, p. 109\,032, Jun. 2023.

\bibitem{liu2024pavedistress}
Z.~Liu \emph{et al.}, ``PaveDistress: A high-resolution pavement distress dataset with fine-grained annotations,'' \emph{Road Mater. Pavement Des.}, pp. 1--19, 2024.

\bibitem{RDD2022}
D.~Arya, H.~Maeda, S.~K. Ghosh, D.~Toshniwal, and Y.~Sekimoto, ``RDD2022: A multi-national image dataset for automatic road damage detection,'' \emph{arXiv preprint arXiv:2209.08538}, 2022.

\bibitem{DSPS}
Y.~Adu-Gyamfi, B.~Buttlar, E.~Dave, D.~Mensching, and H.~Majidifard, ``DSPS: Data science for pavements challenge,'' \emph{[Online]}. Available: \url{https://dsps-1e998.web.app/data}, accessed Feb. 9, 2025.

\bibitem{UAPD}
J.~Zhu \emph{et al.}, ``Pavement distress detection using convolutional neural networks with images captured via UAV,'' \emph{Autom. Constr.}, vol. 133, p. 103991, Mar. 2022.

\bibitem{owor2024pavesam}
N.~J. Owor, Y.~Adu-Gyamfi, A.~Aboah, and M.~Amo-Boateng, ``PaveSAM—Segment anything for pavement distress,'' \emph{Road Mater. Pavement Des.}, pp. 1--25, 2024.

\bibitem{kyem2024advancing}
B.~A. Kyem \emph{et al.}, ``Advancing pavement distress detection in developing countries: A novel deep learning approach with locally-collected datasets,'' \emph{arXiv preprint arXiv:2408.05649}, 2024.

\bibitem{aboah2023real}
A.~Aboah and N.~J. Owor, ``Real-time pavement distress detection using deep learning,'' in \emph{Proc. IEEE Int. Conf. Big Data}, 2023, pp. 1--7.

\bibitem{danyo2025resnet50pci}
Danyo, A., Dontoh, A., \& Aboah, A. (2025). An improved ResNet50 model for predicting pavement condition index (PCI) directly from pavement images. \textit{Road Materials and Pavement Design}, 1–18. https://doi.org/10.1080/14680629.2025.2498632

\bibitem{yolo11_ultralytics}
Ultralytics, ``YOLOv11,'' \emph{[Online]}. Available: \url{https://github.com/ultralytics/ultralytics}, accessed Feb. 9, 2025.

\bibitem{tian2025yolov12attentioncentricrealtimeobject}
S.~Tian \emph{et al.}, ``YOLOv12: Attention-centric real-time object detector,'' \emph{arXiv preprint arXiv:2501.01563}, 2025.

\bibitem{wang2024yolov9}
C.~Y. Wang \emph{et al.}, ``YOLOv9: Learning what you want to learn using programmable gradient information,'' \emph{arXiv preprint arXiv:2402.13616}, 2024.

\bibitem{wang2024yolov10realtimeendtoendobject}
C.~Y. Wang \emph{et al.}, ``YOLOv10: Real-time end-to-end object detection,'' \emph{arXiv preprint arXiv:2405.14458}, 2024.

\bibitem{ren2016fasterrcnnrealtimeobject}
S.~Ren, K.~He, R.~Girshick, and J.~Sun, ``Faster R-CNN: Towards real-time object detection with region proposal networks,'' \emph{IEEE Trans. Pattern Anal. Mach. Intell.}, vol.~39, no.~6, pp. 1137--1149, Jun. 2017.

\bibitem{dontoh2025visualdominanceemergingmultimodal}
A.~Dontoh \emph{et al.}, ``Visual dominance and emerging multimodal approaches in distracted driving detection: A review of machine learning techniques,'' arXiv preprint arXiv:2505.01973, 2025.

\bibitem{carion2020end}
N.~Carion \emph{et al.}, ``End-to-end object detection with transformers,'' in \emph{Proc. Eur. Conf. Comput. Vis. (ECCV)}, 2020, pp. 213--229.

\end{thebibliography}
\end{document}